\documentclass[10pt,times,mathptm,psfig,twocolumn,journals]{IEEEtran}
\usepackage{graphicx}
\usepackage{amsmath}
\usepackage{amssymb}
\usepackage{booktabs}
\usepackage{bm}
\usepackage{mathrsfs}
\usepackage{makecell}
\usepackage{cuted}
\usepackage{capt-of}
\usepackage{pifont}
\usepackage{color}
\usepackage{ulem}
\usepackage{cite}
\usepackage{hyperref}

\newcommand{\xmark}{\ding{55}}
\newcommand{\cmark}{\ding{51}}%

\begin{document}

\title{FisheyeEX: Polar Outpainting for Extending the FoV of Fisheye Lens}

\author{Kang~Liao,~\IEEEmembership{Student Member,~IEEE}, Chunyu~Lin,~\IEEEmembership{Member,~IEEE}, Yunchao Wei,~\IEEEmembership{Member,~IEEE}, Yao~Zhao,~\IEEEmembership{Senior Member,~IEEE}
\thanks{Kang Liao, Chunyu Lin, Yunchao Wei, and Yao Zhao are with the Institute of Information Science, Beijing Jiaotong University, Beijing 100044, China, and also with the Beijing Key Laboratory of Advanced Information Science and Network Technology, Beijing 100044, China (email: kang\_liao@bjtu.edu.cn, cylin@bjtu.edu.cn, wychao1987@gmail.com, yzhao@bjtu.edu.cn). \textit{(Corresponding author: Chunyu Lin)}}
}

\maketitle
\begin{abstract}
Fisheye lens gains increasing applications in computational photography and assisted driving because of its wide field of view (FoV). However, the fisheye image generally contains invalid black regions induced by its imaging model. In this paper, we present a FisheyeEX method that extends the FoV of the fisheye lens by outpainting the invalid regions, improving the integrity of captured scenes. Compared with the rectangle and undistorted image, there are two challenges for fisheye image outpainting: irregular painting regions and distortion synthesis. Observing the radial symmetry of the fisheye image, we first propose a polar outpainting strategy to extrapolate the coherent semantics from the center to the outside region. Such an outpainting manner considers the distribution pattern of radial distortion and the circle boundary, boosting a more reasonable completion direction. For the distortion synthesis, we propose a spiral distortion-aware perception module, in which the learning path keeps consistent with the distortion prior of the fisheye image. Subsequently, a scene revision module rearranges the generated pixels with the estimated distortion to match the fisheye image, thus extending the FoV. In the experiment, we evaluate the proposed FisheyeEX on three popular outdoor datasets: Cityscapes, BDD100k, and KITTI, and one real-world fisheye image dataset. The results demonstrate that our approach significantly outperforms the state-of-the-art methods, gaining around 27\% more content beyond the original fisheye image.
\end{abstract}

\markboth{}
{Shell \MakeLowercase{\textit{et al.}}: Bare Demo of IEEEtran.cls for IEEE Transactions on Magnetics Journals}
\IEEEpeerreviewmaketitle

\section{Introduction}
\IEEEPARstart {F}{ISHEYE} lens has been widely used in computational photography and assisted driving since it can capture a larger field of view (FoV) than the conventional lens. This advantage enables the intelligence system to understand large-scale scenes and conduct timely decision-making. However, the fisheye lens has a special circular surface, leading to the captured scene displayed in a circle construction surrounded by invalid black regions. Having observed this fact, we actively seek a solution to extrapolate the fisheye image and generate a larger FoV than the original fisheye lens.

\begin{figure}[t]
\centering
\includegraphics[width=1\linewidth]{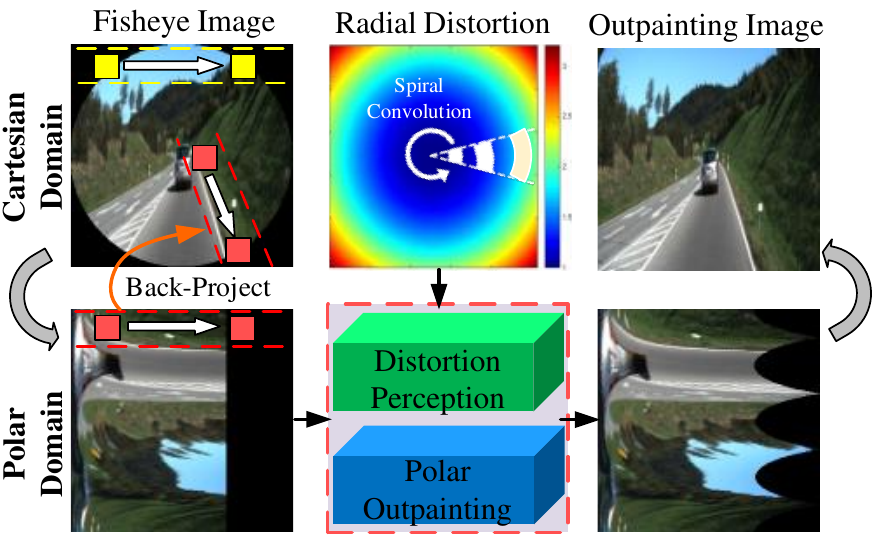}
\caption{Overview of our FisheyeEX method. Given a fisheye image, FisheyeEX extrapolates the semantically coherent details to the black invalid region, improving the integrity of the captured scene with a larger field of view. Compared to the outpainting manner in the Cartesian domain (yellow kernel), our polar outpainting strategy (red kernel) offers a more reasonable completion direction that generates the content from the center to the outside, which also allows a spiral convolution manner to learn the radial distortion in the fisheye image.}
\label{fig:teaser}
\end{figure}

Outpainting, also dubbed image extrapolation, is to generate semantically consistent contents beyond the original boundaries given an image \cite{111, 12, 13, 14, 29}. Compared with the inpainting \cite{15, 151, 16, 17, 18, 181, 19, 20}, outpainting is more challenging due to the one-side constraint. Previous outpainting works usually consider the manual intervention case in which the original image is cropped to a sub-image, and thus it is difficult to apply them into practice. In this work, we extend the FoV of the fisheye lens in the polar domain and extrapolate the invalid regions by fully considering its distortion prior.

Particularly, we propose a FisheyeEX method as illustrated in Figure \ref{fig:teaser}. Given a fisheye image, FisheyeEX extrapolates the semantically coherent content into the black invalid region, improving the integrity of captured scenes with a larger FoV. However, directly implementing the conventional outpainting in fisheye images is challenging: (i) \textit{Outpainting manner}. In contrast to the straight boundary in previous methods, our FisheyeEX extrapolates the fisheye image from the circle boundary in irregular painting regions. In addition, the previous left-to-right completion direction ignores the spatial correlation between the generated content and the original content. (ii) \textit{Distortion synthesis}. Unlike the natural image, the radial distortion exists in the fisheye image, which allows larger FoV for the captured scene. Therefore, besides the content generation, FisheyeEX needs to estimate and synthesize the radial distortion during the outpainting process.

To address the above challenges, we present a novel polar outpainting strategy and a feasible learning framework. Considering the radial symmetry of the fisheye image, we first transform the outpainting from the Cartesian domain into the polar domain. As a result, the boundary of the fisheye image is converted from a circle into straight lines and the discrete irregular filling regions are gathered together. More importantly, the polar outpainting enables a more reasonable extrapolation direction, which generates the coherent semantics from the center to the boundary. For the distortion synthesis, a spiral distortion-aware perception module is used to estimate the original distortion, and then a scene revision module rearranges the generated content with the predicted radial distortion. In the experiment, we evaluate the proposed FisheyeEX on three popular datasets: Cityscapes, BDD100k, and KITTI, and one real-world fisheye image dataset. The results demonstrate that our approach significantly outperforms the state-of-the-art image completion methods, gaining around 27\% more content beyond the original fisheye image. 

In summary, our contributions can be listed as follows:

\begin{itemize}
    \item We propose a practical FisheyeEX method to extend the Fov of fisheye lens, in which a polar outpainting strategy enables a reasonable completion direction based on the radial symmetry of the fisheye image.
    \item Besides the content generation, a spiral distortion-aware perception module and a scene revision module are designed to guide the distortion synthesis of the outpainting results.
\end{itemize}

The rest of this paper is organized as follows. We first introduce the related work in Section \ref{sec2}. Subsequently, we present our polar outpainting framework in Section \ref{sec3}. The experiments are provided in Section \ref{sec4}. Finally, we conclude this paper in Section \ref{sec5}. 

\section{Related Works}
\label{sec2}

In this section, we briefly review the previous works regarding to the image outpainting and distortion estimation for fisheye image.

\noindent \textbf{Image Outpainting}
Due to the one-side constraint, image outpainting is more challenging than image inpainting, which can be fairly classified into traditional methods \cite{25, 26, 27} and learning methods \cite{111, 12, 13, 14, 29}. In analogy to inpainting, traditional outpainting methods \cite{25, 26, 27} share a similar pipeline where the relevant patches are retrieved in the candidate set, and the matched patches are merged into the original input. Inspired by the powerful generation ability of the generative adversarial networks (GANs) \cite{28}, Sabini et al. \cite{111} first realized the parametric outpainting with the three-phase training procedure. Teterwak et al. \cite{12} conditioned the discriminator with pre-trained features, promoting the generated results to match the target image semantically. Wang et al. \cite{13} introduced a feature expansion module and content prediction module to relieve the difficulty of one-pass outpainting. Yang et al. \cite{14} explored a very long natural scenery image prediction by outpainting, and it extrapolated the natural scenery images horizontally in multi-steps. Motivated by humans' perception fashion, Guo et al. \cite{29} designed a novel slice-wise GAN to extrapolate the unseen content outside the boundary of the image.

Inspired by the promising efforts contributed by the above methods, we explore to design a practical outpainting framework to enlarge the FoV of the fisheye image.

\begin{figure*}[t]
\centering
\includegraphics[width=1\linewidth]{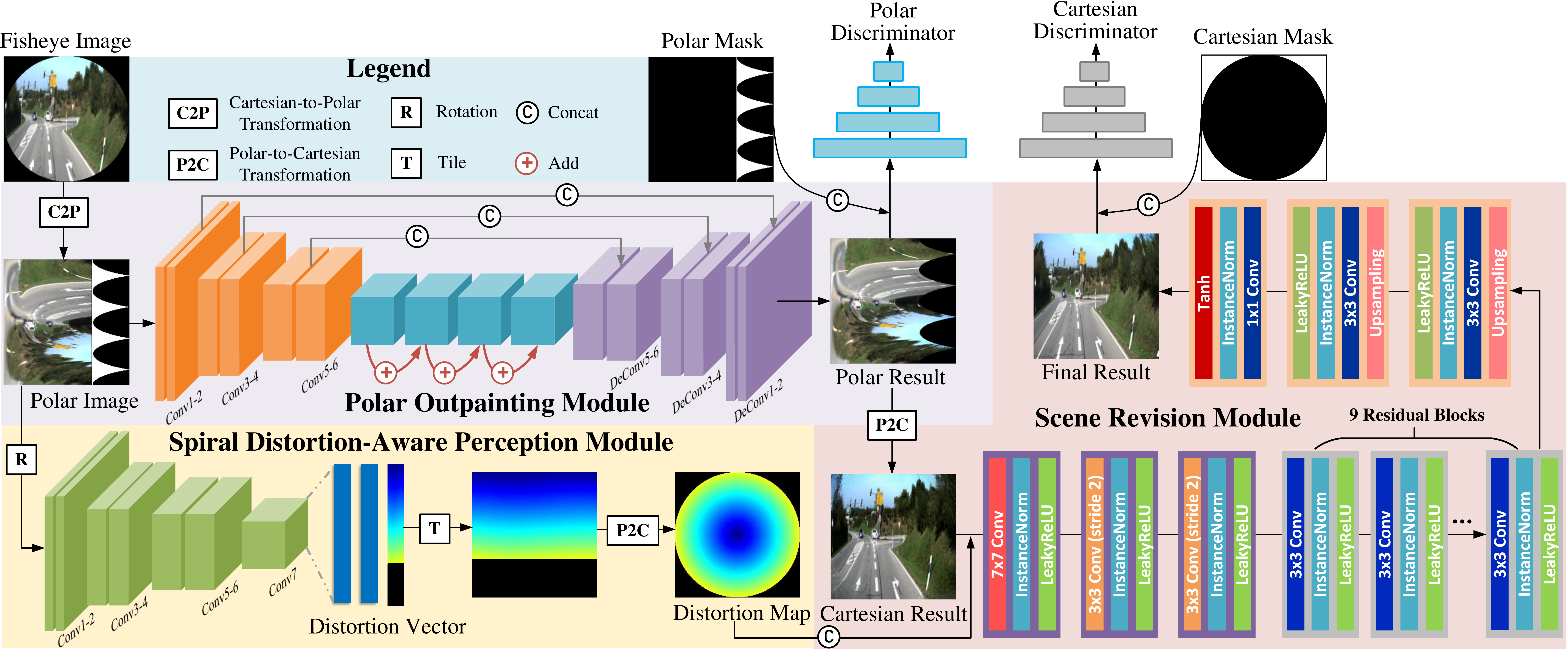}
\caption{Overview of our network architecture with polar outpainting module, spiral distortion-aware perception module, and scene revision module.}
\label{fig:network}
\end{figure*}

\noindent \textbf{Distortion Estimation}
In contrast to the natural image, the fisheye image captured by the fisheye lens contains radial distortion. This distortion changes the original geometry shape of objects, thus enabling a larger field of view for the captured scene. The farther pixel to the optical center, the stronger the distortion is, and vice versa. As for some computer vision tasks such as object detection, it is crucial to build an accurate distortion estimation for the fisheye image. For example, the traditional methods \cite{1, 2, 3, 4, 5} estimated the distortion parameter of the distortion image based on the hand-crafted features and usually required manual intervention. To achieve more robust and flexible algorithms, the learning methods \cite{6, 7, 8, 10, 11, 43} rectified the distortion image in terms of the deep semantic features. In particular, Liao et al. \cite{8} predicted a 2D distortion distribution map (DDM) to generate explicit guidance for networks. Similarly, Li et al. \cite{11} constructed a 2D displacement flow using an encoder-decoder architecture to help the distortion rectification. \cite{50} predicted the displacement flow and rectified the fisheye image in the polar coordinates, which shows superior rectification performance. In contrast to the above distortion rectification methods, our aim is to perceive an explicit distortion level and \textit{synthesize} the radial distortion for more realistic fisheye image outpainting. 
However, most works omitted the fisheye image's radial symmetry and directly leveraged the learning strategy from the natural image to extract the distortion features. 

In this work, invalid black regions are required to be filled to enlarge the FoV of fisheye image. To keep the extrapolated regions matching the original fisheye image, we have to consider the distortion synthesis, which is challenging and different from the classical outpainting. By converting the fisheye image into the polar domain, we can achieve a spiral convolution manner to learn the accurate distribution of radial distortions and achieve the distortion synthesis during the outpainting.

\section{Our Method}
\label{sec3}
We propose a novel polar outpainting strategy for the fisheye image as shown in Figure \ref{fig:network}, transforming the outpainting reference system from the Cartesian domain into the polar domain. Given an input fisheye image $I_f \in \mathbb{R}^{h\times w\times 3}$ and the filling mask $M \in \mathbb{R}^{h\times w\times 1}$, the goal of our FisheyeEX is to extrapolate the original content and details to the outside black region, improving the integrity of captured scenes with the larger field of view (FoV). 

To be specific, our FisheyeEX consists of three modules: the polar outpainting module, the spiral distortion-aware perception module, and the scene revision module. 

\subsection{Polar Outpainting Module}
\label{s3.1}
\noindent \textbf{Polar Coordinates Transformation}
Due to the fisheye lens's special circular surface, the captured scene is displaced in a circle construction, and the induced radial distortion exists in the fisheye image. In general, the radial distortion is considered centrally symmetric with the optical center. The distortion degree of a pixel is positive to its distance to the symmetric center. Having observed this fact, we transform the outpainting reference system from the conventional Cartesian domain into the polar domain, reconstructing the geometry structure and the distortion distribution of the fisheye image. 

In particular, we suppose that the point $\mathbf{P}_c = (x, y)$ and the point $\mathbf{P}_p = (\rho, \theta)$ locate at the Cartesian coordinates and the polar coordinates, respectively. First, we choose the optical center $\mathbf{P}_o = (x_{o}, y_{o})$ as the polar point. Then, the polar coordinates can be established counterclockwise by rotating a ray of $\mathbf{P}_o$. The coordinates transformation relationship is expressed as:

\begin{equation}\label{1}
\begin{aligned}
\rho& =\sqrt{(x-x_{o})^{2}+(y-y_{o})^{2}}\\
\theta& =\arctan(\frac{y-y_{o}}{x-x_{o}}), 
\end{aligned}
\end{equation}

\noindent where $\rho$ denotes the distance from the pixel to the optical center and $\theta$ is the corresponding angle.

In the fisheye image outpainting case, we can offer the following advantages by transforming the outpainting from the Cartesian domain into the polar domain.

1. The extrapolation direction is more reasonable and effective. Previous outpainting methods directly use the completion manner in image inpainting. However, the plain left-to-right direction ignores the challenging spatial correlation of generated content in the outpainting case. Instead, by reconstructing the fisheye image to the polar domain, our approach can build an effective inside-to-outside direction for better outpainting, as illustrated in Fig. \ref{fig:teaser}.

\begin{figure}[t]
\centering
\includegraphics[width=1\linewidth]{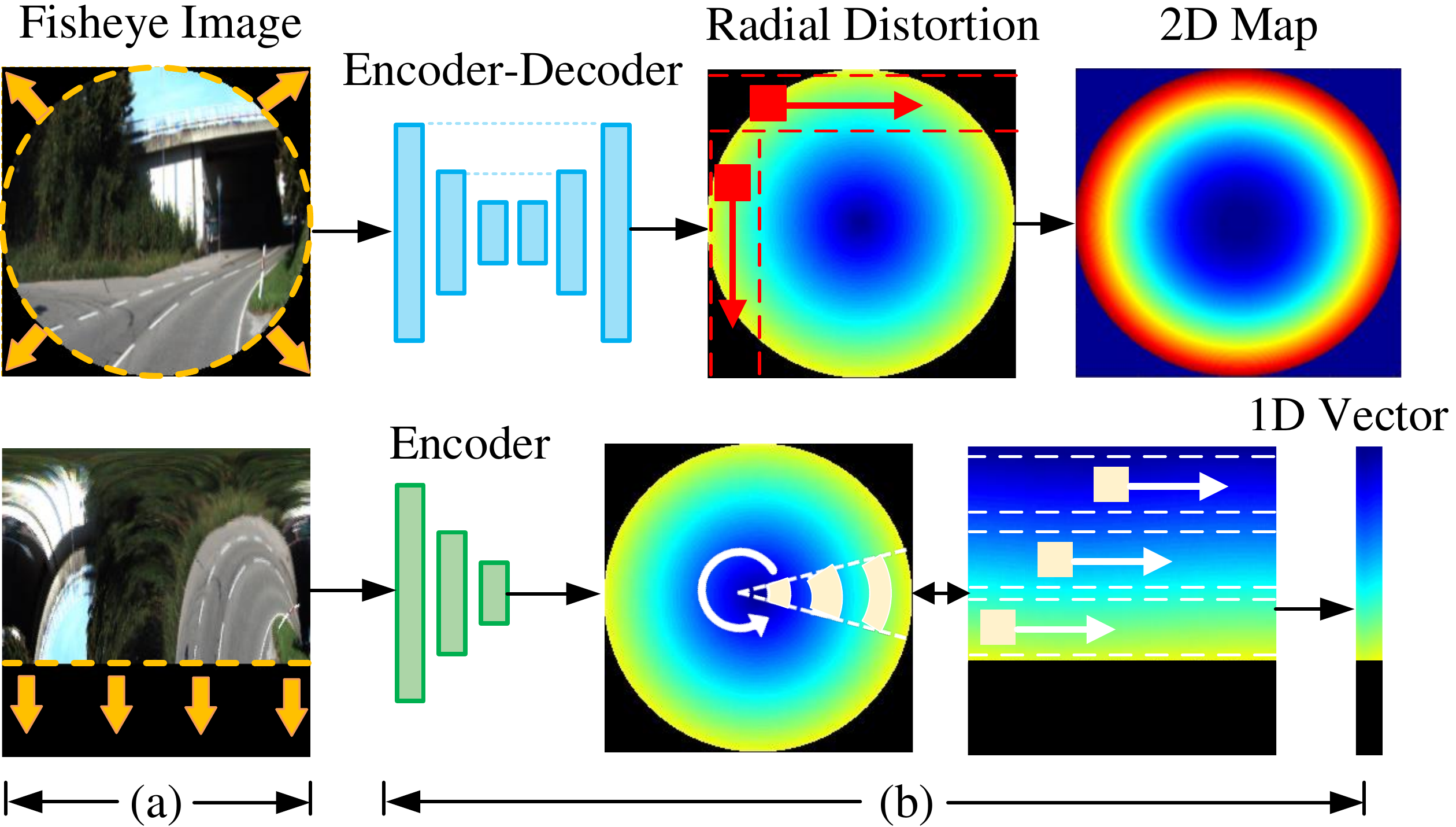}
\caption{Advantages of the polar domain for the fisheye image. (a) The circle boundary of the fisheye image is reconstructed into the straight boundary. (b) The learning path of convolutional kernels keeps consistent with the distortion distribution of the fisheye image, which is back-projected as a spiral learning manner in the Cartesian domain. Additionally, compared with previous methods, our spiral distortion-aware perception method is more efficient due to the lightweight network and low dimensional target.}
\label{fig:polar}
\end{figure}

2. The circle boundary of the fisheye image and the discrete painting region are converted into a straight and continuous region, as shown in Fig. \ref{fig:polar} (a). Thus, the simplified structure can relieve the outpainting's difficulty and improve the extrapolation performance.

3. The distortion perception is more accurate and efficient. Different from the natural image, the fisheye image has radial distortion. As shown in Fig. \ref{fig:polar} (b), in the polar domain, the radial distortion is rearranged, and the learning path of convolutional kernels keeps consistent with the new distortion distribution, facilitating more accurate distortion perception. Moreover, we only need to predict a 1D vector of distortion values while previous methods require a more computational network to predict 2D distortion maps \cite{8, 11}.

\noindent \textbf{Fisheye Image Extrapolation}
After the domain transformation, the original fisheye image $I_f \in \mathbb{R}^{h\times w\times 3}$ is transformed into the polar fisheye image $I_f^p \in \mathbb{R}^{h'\times w'\times 3}$. Then, we feed $I_f^p$ and mask into the outpainting module to conduct the content generation in the filling regions.

The outpainting module can be regarded as a classical fully convolutional encoder-decoder network. In the encoder, there are 3 hierarchies progressively extracting the semantic features of the polar fisheye image, where each hierarchy includes two convolutional layers with the LeakyReLU activation function. After each hierarchy except the last one, the resolution of the feature map is down-sampled. Subsequently, a residual convolutional group with the dilation rate of 2, 4, 8, 16, captures the long-range spatial information with a large receptive field in the encoded features. Receiving the high dimension semantic features from the encoder, the decoder network generates realistic content beyond the reconstructed straight boundary. At the beginning of each hierarchy in the decoder, a bilinear upsampling layer (×2) is used to increase the spatial dimension of the feature map. Each convolutional layer is activated by the LeakyReLU function in the decoder, except that the activation of the last convolutional layer is the Tanh function clipping its outputs to the range $[-1, 1]$. To introduce the low-level features with rich texture, a skip connection operation is implemented on the same resolution feature maps from the encoder to the decoder. 

\subsection{Spiral Distortion-Aware Perception Module}
\label{s3.2}
Compared with the natural image, the fisheye image contains radial distortion to provide a large FoV. Therefore, besides the content generation, we need to synthesize the radial distortion in the fisheye image outpainting task. The first step is to perceive and estimate the original distortion in fisheye image. Suppose that a pixel in the natural image is $\mathbf{P}_o = (x_n, y_n)^{\rm T} \in {\mathbb{R}}^{2\times1}$ and its corresponding pixel in the fisheye image is $\mathbf{P}_f = (x_f, y_f)^{\rm T} \in {\mathbb{R}}^{2\times1}$, then the transformation of coordinates can be described by
\begin{equation}\label{eq1-c}
\begin{split}
& x_f = x_n(1 + {k_1}{{r}^2_f} + {k_2}{{r}^4_f} + {k_3}{{r}^6_f} + {k_4}{{r}^8_f} + \cdots) \\
& y_f = y_n(1 + {k_1}{{r}^2_f} + {k_2}{{r}^4_f} + {k_3}{{r}^6_f} + {k_4}{{r}^8_f} + \cdots), \\
\end{split}
\end{equation}
where $[k_1, k_2, k_3, k_4, \cdots]$ are the radial distortion parameters and $r_f$ indicates the Euclidean distance between the distorted pixel and the distortion center $\mathbf{P}_c = (x_c, y_c)^{\rm T} \in {\mathbb{R}}^{2\times1}$ in the fisheye image. Like DDM \cite{8}, we offer an explicit learning representation: distortion level $\mathcal{D}$, for the distortion perception, which shows more intuitive visual relations to the fisheye image than the distortion parameters. For a pixel $\mathbf{P}_f$ in the fisheye image, its distortion level $\mathcal{D}(x_f, y_f)$ can be computed by building a ratio equation based on the Eq. \ref{eq1-c}:
\begin{equation}\label{eq4}
\mathcal{D}(x_f, y_f) = \frac{x_n}{x_f} = \frac{y_n}{y_f} = 1 + {k_1}{{r}^2_f} + {k_2}{{r}^4_f} + {k_3}{{r}^6_f} + \cdots.
\end{equation}

To be more specific, we design a spiral distortion-aware perception module consisting of an encoder network and a header network, to guide the distortion synthesis of outpainting results. The encoder network is to extract the general representation of the distortion context in the form of the high-level semantic features using stacked convolutional layers. During the feature extraction in polar domain, our approach enables a consistent convolutional learning path with the distortion distribution: the kernels at the same row learn the same distortion level. This learning manner is back-projected as a spiral convolution in the Cartesian domain, following the radial symmetry of the fisheye image. In contrast, the conventional convolution omits the crucial distribution rule in the fisheye image. The header network, composed of fully connected layers, combines the general representation of the distortion feature and constructs a 1D feature vector to estimate the distortion level. As a result, we can achieve more accurate distortion perception by learning the radial distortion in the polar domain. Moreover, our learning manner is more efficient since we only need to predict a 1D distortion vector rather than a 2D distortion map.

\begin{figure}[t]
\centering
\includegraphics[width=1\linewidth]{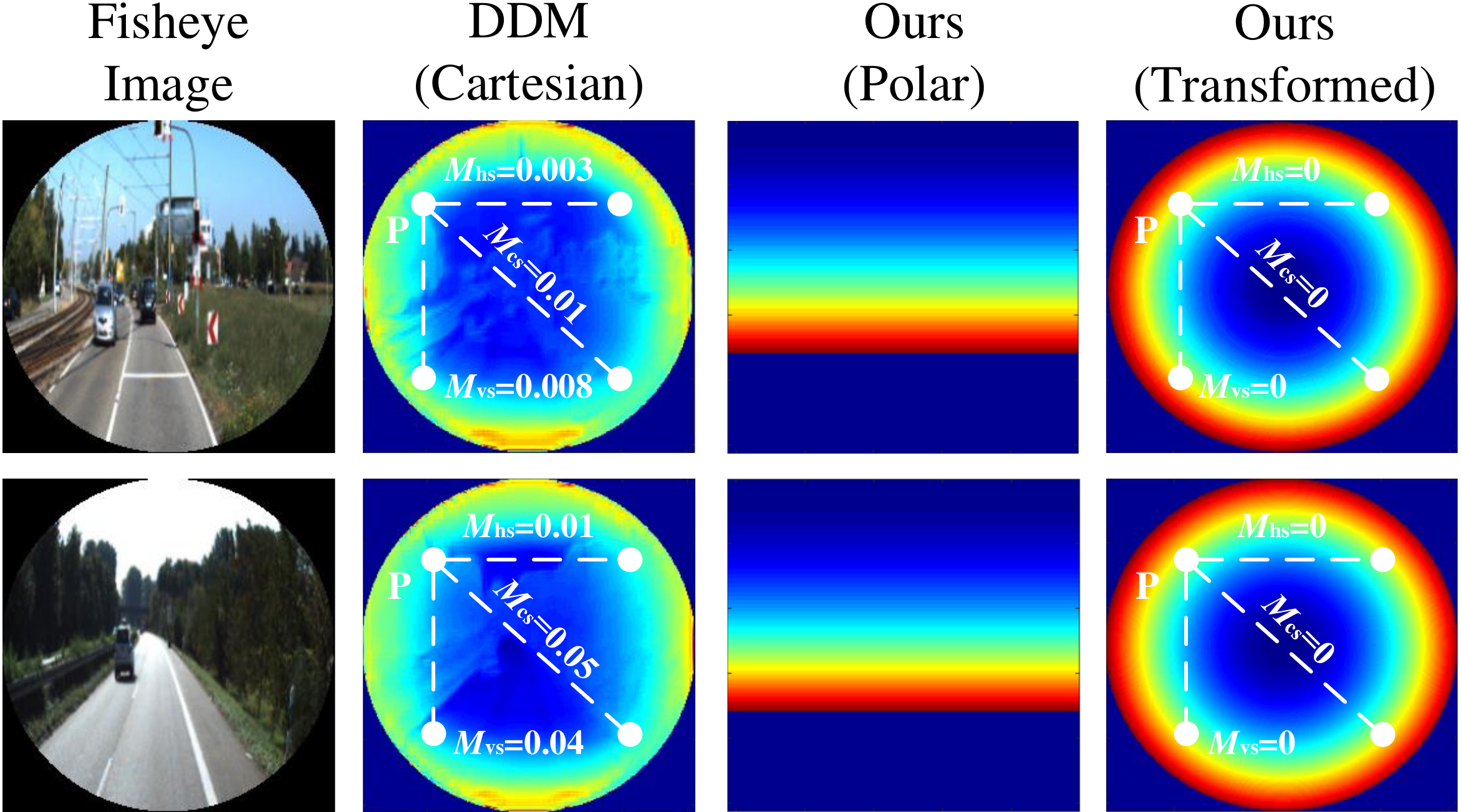}
\caption{Visual comparison of the Cartesian distortion perception \cite{8} and our polar distortion perception. For each result displayed in Cartesian domain, we show a pixel $P$ marked with $\mathcal{M}_{hs}$, $\mathcal{M}_{vs}$, and $\mathcal{M}_{cs}$ for the radial symmetry evaluation. Compared with DDM \cite{8}, our method can estimate more accurate distortions of fisheye images and our distortion distributions follow the strict radial symmetry, where $\mathcal{M}_{hs}=\mathcal{M}_{vs}=\mathcal{M}_{cs}=0$.}
\label{fig:ddm}
\end{figure}

For the evaluation, besides the $\mathcal{L}_1$ error between the estimated distortion value and the ground truth, we further present three evaluation metrics based on the radial symmetry of the fisheye image: central symmetry metric $\mathcal{M}_{cs}$, horizontal symmetry metric $\mathcal{M}_{hs}$, and vertical symmetry metric $\mathcal{M}_{vs}$. Supposed $\hat \delta(x_i, y_i)$ is an estimated distortion level, then $\mathcal{M}_{hs}$, $\mathcal{M}_{vs}$, and $\mathcal{M}_{cs}$ can be defined as follows.

\begin{equation}\label{eq-radial}
\begin{split}
\mathcal{M}_{hs} = \frac{1}{WH}\sum_{i=0}^{W-1}\sum_{j=0}^{H-1}|| \hat \delta(x_i, y_i) - \hat \delta(\frac{W}{2} - x_i + x_c, y_i)||_1,\\
\mathcal{M}_{vs} = \frac{1}{WH}\sum_{i=0}^{W-1}\sum_{j=0}^{H-1}|| \hat \delta(x_i, y_i) - \hat \delta(x_i, \frac{H}{2} - y_i + y_c)||_1,
\end{split}
\end{equation}

\begin{equation}\label{eq-radial}
\begin{split}
\mathcal{M}_{cs} =
\frac{1}{WH}\sum_{i=0}^{W-1}\sum_{j=0}^{H-1}|| \hat \delta(x_i, y_i)- & \hat \delta(\frac{W}{2} - x_i + x_c,\\
&\frac{H}{2} - y_i + y_c)||_1,\\
\end{split}
\end{equation}
where $(x_c, y_c)$ indicates the optical center of the fisheye image. Figure \ref{fig:ddm} illustrates the visual comparison of the Cartesian distortion perception \cite{8} and polar distortion perception results. The distortion map transformed by our polar perception result exhibits a clear and strict radial symmetry, while the result generated by DDM \cite{8} shows a disordered and inaccurate distortion distribution. The quantitative evaluation and network complexity comparison will be demonstrated in Section \ref{s4.3}.

\subsection{Scene Revision Module}
Since most images are shown in standard Cartesian coordinates, we need to recover the extrapolation results from the polar domain to the Cartesian domain. Moreover, the outpainting results should follow the distribution of the radial distortion. After the distortion perception, the scene revision module takes the polar outpainting result and the estimated distortion as inputs to rearrange the generated pixel with the geometric distortions, shown in Figure \ref{fig:network}. First, the transformed extrapolated image and distortion map are fed into a base feature aggregation group consisting of three convolutional layers with the LeakyReLU activation function. The last two convolutional layers are performed using stride $2$. Then, a deep residual group takes the feature maps with the size of $\frac{h}{4}\times \frac{w}{4} \times 256$ as input, gradually extracting the high-level semantic features with nine residual blocks. Each residual block comprises a convolutional layer activated using the ReLU function and an instance normalization layer. Finally, two deconvolutional layers upsample the residual feature maps and generate the refine extrapolated results. To boost high-quality image generation, we introduce the adversarial learning strategy in the framework. In detail, a polar discriminator and a Cartesian discriminator are used to judge if the generated result is real or fake with the guidance of a polar mask and Cartesian mask, facilitating a region-aware competition with the generator network.

\subsection{Training}
\noindent \textbf{Loss Function}
For the polar outpainting module, we train it using a perceptual reconstruction loss $\mathcal{L}_{pr}$ and an adversarial loss $\mathcal{L}_{ad}$. $\mathcal{L}_{pr}$ computes the pixel-wise difference between generated outpainting fisheye image $\hat I_{op}$ and ground truth $I_{op}$ on the feature maps $\phi_{i,j}$, which are obtained from the $j$-th convolution (after activation) before the $i$-th maxpooling layer in the VGG19 network \cite{36}:
\begin{equation}\label{loss1}
\begin{split}
\mathcal{L}_{pr} = \frac{1}{W_{i,j}H_{i,j}} & \sum_{x=1}^{W_{i,j}}\sum_{y=1}^{H_{i,j}}||\phi_{i,j}{( I_{op})}_{x,y} - \phi_{i,j}{(\hat I_{op})}_{x,y}||_2.
\end{split}
\end{equation}

$\mathcal{L}_{ad}$ indicates the wasserstein-GAN (WGAN) loss \cite{433}, it can improve the stability of adversarial training and the quality of generated images. The WGAN value function can be described by the Kantorovich-Rubinstein duality \cite{60}:

\begin{equation}\label{eq_wgan}
\min_G \max_{C \in \mathcal{C}}  \mathop{\mathbb{E}}\limits_{\bm{x} \sim \mathbb{P}_r} [C(\bm{x})] - \mathop{\mathbb{E}}\limits_{\tilde{\bm{x}} \sim \mathbb{P}_g}[C(\tilde{\bm{x}})];
\end{equation}
where $G$ and $C$ are the generator network and critic network in WGAN. $\mathcal{C}$ is the set of 1-Lipschitz functions, and $\mathbb{P}_g$ is once again the model distribution. Given an input $X$ to the generator network, the formulation of $\mathcal{L}_{ad}$ is constructed as follows:

\begin{equation}\label{loss_was}
\mathcal{L}_{ad} = -\sum_{n = 1}^{N} C_{\theta_C}(G_{\theta_G}(X)).
\end{equation}

For the spiral distortion-aware perception module, we implement the $\mathcal{L}_{1}$ loss to learn its parameters. Suppose the estimated distortion and ground truth are $\mathcal{\hat D}$ and $\mathcal{D}$, respectively. Then, the loss function can be formed by

\begin{equation}\label{loss_dis}
\mathcal{L}_{sd} = \frac{1}{N}\sum_{i=1}^N||\mathcal{D}_i - \mathcal{\hat D}_i||_1.
\end{equation}

During the training process, we first jointly train the polar outpainting module and spiral distortion-aware perception module, the total loss $\mathcal{L}_{t}$ can be represented by

\begin{equation}\label{loss_total}
\mathcal{L}_{t} = \mathcal{L}_{pr} + \lambda_{ad}\mathcal{L}_{ad} + \lambda_{sd}\mathcal{L}_{sd},
\end{equation}
where we empirically set $\lambda_{ad} = 0.05$ and $\lambda_{sd} = 0.1$. In analogy to the polar outpainting module, we then train the scene revision module with the perceptual reconstruction loss and the adversarial loss. The main difference is that we use the shallower feature maps in VGG19: $\phi_{1,2}, \phi_{2,2}, \phi_{3,3}$, to boost a fine details enhancement.

\noindent \textbf{Implementation Details}
We first train the polar outpainting module and spiral distortion-aware perception module to extrapolate the fisheye image and estimate its radial distortion, optimized using Adam \cite{46} with the learning rate of $1\times10^{-3}$ and the parameters $\beta_1 = 0.5$ and $\beta_2 = 0.9$. Subsequently, we train the scene revision module using Adam with the learning rate of $5\times10^{-4}$, to rearrange the generated pixel with the estimated distortion distribution. All networks are trained on NVIDIA GeForce RTX 2080 Ti GPUs, and the mini-batch size is 4.

\begin{figure}[t]
\centering
\includegraphics[width=1\linewidth]{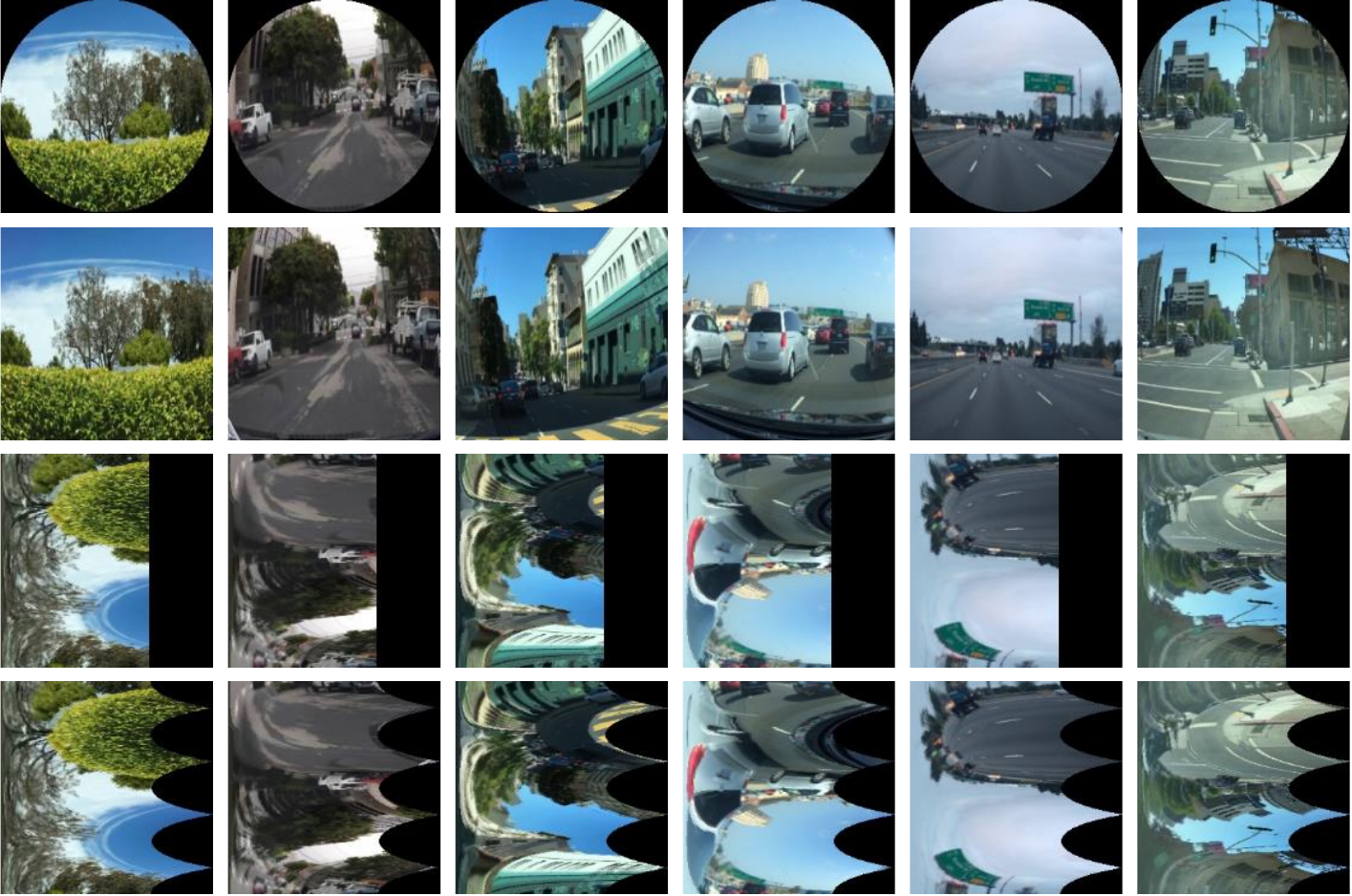}
\caption{Samples of our fisheye image outpainting dataset. We show the fisheye image, extrapolated image, polar fisheye image, and polar extrapolated image, from top to bottom.}
\label{fig:dataset}
\end{figure}

\section{Experiments}
\label{sec4}
In this section, we first describe the details of the fisheye image datasets including three synthesized datasets and one real-world dataset. Subsequently, the experimental results of our polar outpainting compared with the state-of-the-art methods are demonstrated, in quantitative measurement, visual qualitative appearance, and user study. Additionally, we discuss the benefits of the polar outpainting compared with the previous Cartesian outpainting. To demonstrate the effectiveness of each module in our framework, we conduct an ablation study to show the different performances. 

\subsection{Dataset}
To evaluate the performance of our FisheyeEX, we conduct experiments on three popular outdoor datasets: Cityscapes \cite{44}, BDD100k \cite{45}, and KITTI \cite{47}. Then, we construct the radial distortion for each image based on Eq. \ref{eq1-c}, following the principles of the parameter selection described in previous methods \cite{8, 11}. To be specific, we leverage the fourth order polynomial model to synthesize the fisheye image, which is sufficient for the approximation of most application scenarios. Then, four distortion parameters are randomly generated from the following ranges: $k_1 \in [-1\times10^{-4}, -1\times10^{-8}]$, $k_2 \in [1\times10^{-12}, 1\times10^{-8}]$ or $\in [-1\times10^{-8}, -1\times10^{-12}]$, $k_3 \in [1\times10^{-16}, 1\times10^{-12}]$ or $\in [-1\times10^{-12}, -1\times10^{-16}]$, and $k_4 \in [1\times10^{-20}, 1\times10^{-16}]$ or $\in [-1\times10^{-16}, -1\times10^{-20}]$. Subsequently, we transform the fisheye image and outpainting ground truth into the polar domain. Thus, our dataset contains four types of data: fisheye image, extrapolated image, polar fisheye image, and polar extrapolated image. As illustrated in Figure \ref{fig:dataset}, we show six samples of the established fisheye image dataset.

Moreover, we also validate the generalization ability of different methods on the real-world fisheye image dataset LMS \cite{48}, in which the Fujinon FE185C057HA-1 fisheye lens was used to capture various scenes. The details of datasets are listed in Table \ref{tab:dataset}.

\begin{table}[t]\small
  \caption{The detailed training and testing split on four datasets.}
  \label{tab:dataset}
	 \centering
   \begin{tabular}{l|c|c|c|c}
    \hline
    Dataset & Type & \#Train & \#Test & \#Total \\
    \hline
    \hline
    
    Cityscapes \cite{44} & Synthesized & 4500 & 500 & 5000\\
    BDD100k \cite{45} & Synthesized & 9000 & 1000 & 10000\\
    KITTI \cite{47} & Synthesized & 14719 & 886 & 15605\\
    \hline
    \hline
    LMS \cite{48} & Real-World & - & 200 & 200\\
  \hline
\end{tabular}
\end{table}

\begin{table*}
\begin{center}
\caption{Quantitative evaluation of the extrapolation results of the state-of-the-art methods. \textbf{\textcolor{red}{Red}} text indicates the best and \textcolor{blue}{blue} text indicates the second-best performing method. We evaluate these methods on three dataset: Cityscapes \cite{44}, BDD100k \cite{45}, and KITTI \cite{47}.}
\label{table:comparison}
\begin{tabular}{l|ccc||ccc||ccc}

\hline
 Dataset &  &  \textit{Cityscapes} &  &  &  \textit{BDD100k} &  &  & \textit{KITTI} &  \\ 
\hline
\hline
Methods & PSNR $\uparrow$ & SSIM $\uparrow$ & FID $\downarrow$ & PSNR $\uparrow$ & SSIM $\uparrow$ & FID $\downarrow$ & PSNR $\uparrow$ & SSIM $\uparrow$ & FID $\downarrow$\\
\hline
SRN \cite{13} & 17.71 & 0.81 & 169.42 & 20.56 & 0.84 & 160.02 & 18.25 & 0.79 & 143.90\\
\hline
RK  \cite{17} & 21.79 & 0.87 & 136.66 & 20.66 & 0.84 & 113.60 & 20.13 & 0.82 & 102.77\\
\hline
HiFill \cite{16} & 22.27 & 0.89 & 109.89 & 21.34 & 0.87 & 98.37 & 20.10 & 0.83 & 82.61\\
\hline
Boundless \cite{12} & {\textcolor{blue}{23.54}} & {\textcolor{blue}{0.90}} & {\textcolor{blue}{64.26}} & {\textcolor{blue}{22.80}} & {\textcolor{blue}{0.88}} & {\textcolor{blue}{84.83}} & {\textcolor{blue}{21.52}}& {\textcolor{blue}{0.87}} & {\textcolor{blue}{53.17}}\\
\hline
Ours & \textbf{\textcolor{red}{24.63}} & \textbf{\textcolor{red}{0.92}} & \textbf{\textcolor{red}{40.06}} &  \textbf{\textcolor{red}{23.14}} & \textbf{\textcolor{red}{0.93}} & \textbf{\textcolor{red}{48.03}} & \textbf{\textcolor{red}{22.31}} & \textbf{\textcolor{red}{0.90}}&  \textbf{\textcolor{red}{34.68}}\\
\hline
\end{tabular}

\end{center}
\end{table*}

\begin{figure*}[t]
\centering
\includegraphics[width=1\linewidth]{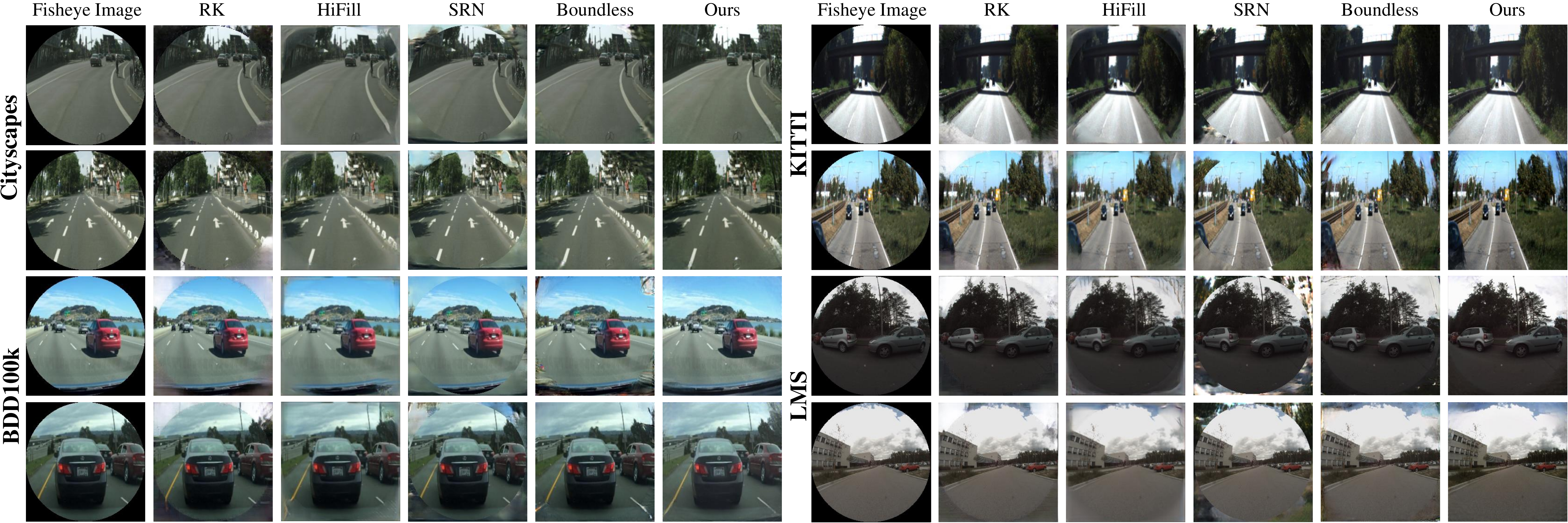}
\caption{Qualitative comparison on different methods. For each comparison, we show the fisheye image, the results derived from RK \cite{17}, HiFill \cite{16}, SRN \cite{13}, Boundless \cite{12}, and Ours.}
\label{fig:cp_sota}
\end{figure*}

\begin{figure*}[t]
\centering
\includegraphics[width=1\linewidth]{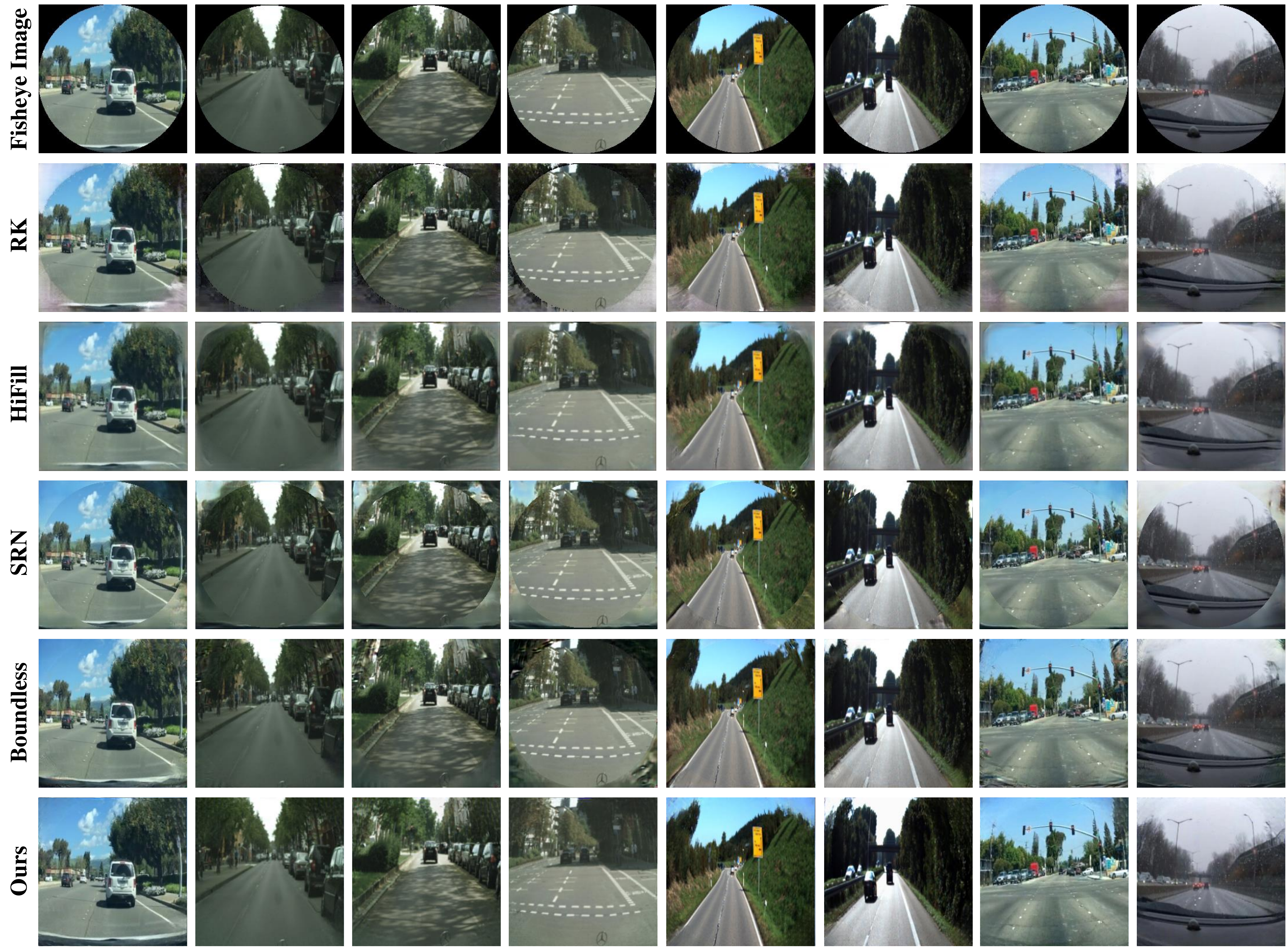}
\caption{More qualitative comparison results of different methods. For each comparison, we show the fisheye image, the results derived from RK \cite{17}, HiFill \cite{16}, SRN \cite{13}, Boundless \cite{12}, and Ours, from top to bottom.}
\label{fig:more_res}
\end{figure*}

\begin{figure}[t]
\centering
\includegraphics[width=1\linewidth]{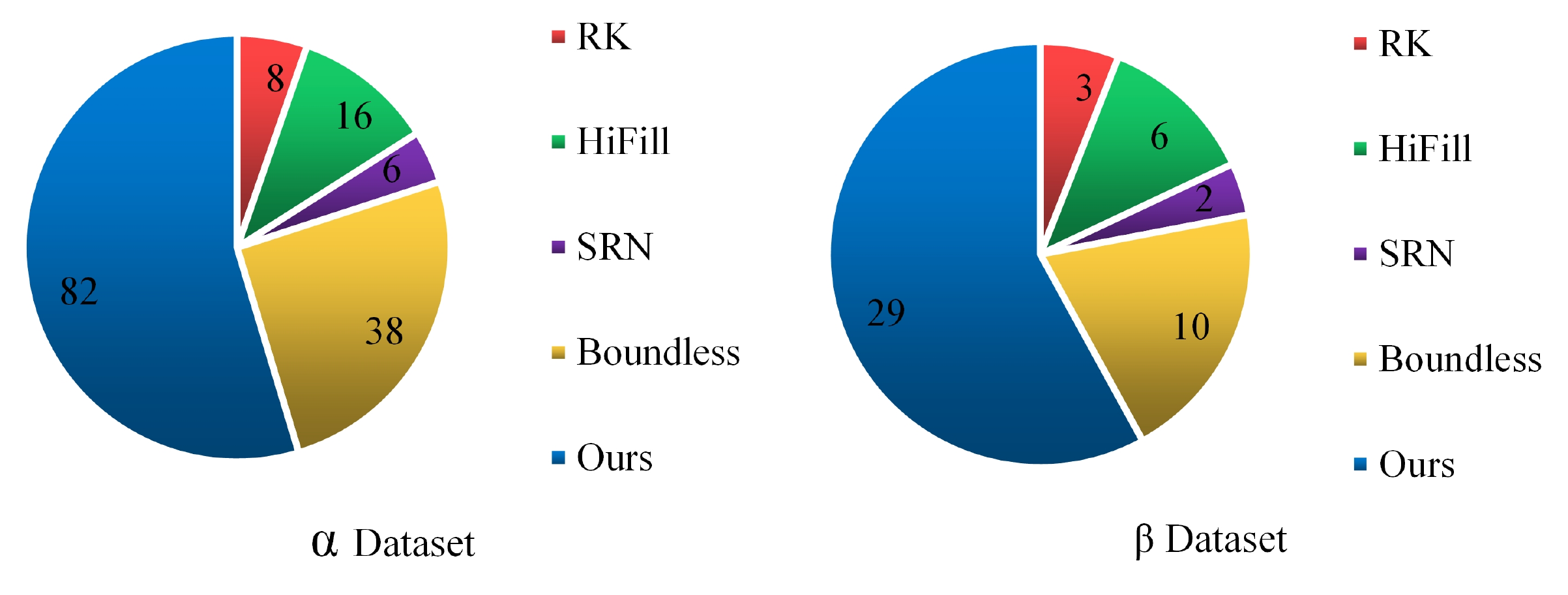}
\caption{User study. We construct the $\alpha$ group including three synthesized dataset and $\beta$ group including a real-word dataset for 20 volunteers to rank different extrapolated images.}
\label{fig:user_study}
\end{figure}

\subsection{Comparison Results}

To comprehensively demonstrate the performance  of our FisheyeEX method, we compare our method with the state-of-the-art image completion methods, including the image inpainting methods: RK \cite{17}, HiFill \cite{16} and image outpainting methods: SRN \cite{13} and Boundless \cite{12}. However, we found that some previous outpainting methods are hard to be applied in the case of the fisheye image extrapolation. For example, Boundless \cite{12} and SRN \cite{13} require the filling region to be a regular rectangle, resulting in inferior and seamed extrapolation results in fisheye images. To show a fair comparison, we retrain Boundless \cite{12} and SRN \cite{13} using our fisheye image dataset with a circle filling region. For the image inpainting methods, RK \cite{17} and HiFill \cite{16} can complete the image with irregular masks flexibly, and thus we directly use the pre-trained models to generate the extrapolated fisheye image.

\noindent \textbf{Quantitative Evaluation:}
Following previous methods, we utilize the peak signal-to-noise ratio (PSNR), structural similarity index measure (SSIM), and Fr\'echet Inception Distance (FID) as metrics for evaluating the semantic consistency and visual realism of generated results. Note that the higher is better for PSNR and SSIM, lower is better for FID. All methods are used to conduct the image completion on three test datasets: Cityscapes, BDD100k, and KITTI. We compute PSNR and SSIM metrics using the pixel difference between each generated image and the ground truth image. For the FID, we measure a Wasserstein-2 distance between the feature space representations of generated and ground truth images using a pre-trained Inception-V3 model \cite{49}. The evaluation results are listed in Table \ref{table:comparison}, showing our method outperforms the state-of-the-art methods in all cases. These improvements are benefited from the proposed polar outpainting strategy, which enables a more reasonable extrapolation manner with a center-to-outside painting direction. Moreover, the designed spiral distortion-aware perception module provides a more accurate distortion distribution for the distortion synthesis.

\noindent \textbf{Qualitative Evaluation:}
Figure \ref{fig:cp_sota} illustrates the qualitative evaluation. The filling region or black valid region of the fisheye image locates between its circle boundary and bounding rectangle. As shown in Figure \ref{fig:cp_sota}, we can observe that the results generated by previous methods usually suffer from artifacts and unsatisfactory details. For example, the results of RK \cite{17} and HiFill \cite{16} occur the fracture of contents especially in the fisheye image boundary. The main reason is that the image inpainting manner ignores the special spatial correlation in the extrapolation case, where the one-side constraint provides very limited information for the content generation. For the image outpainting methods, SRN \cite{13} and Boundless \cite{12} generate some geometrically distorted objects such as the lane line and trees, which is caused by the vanilla left-to-right outpainting manner. In contrast, as a benefit of the proposed polar outpainting, we can achieve a more reasonable outpainting manner that extrapolates the semantically consistent details from the image center to the outside. As a result, our FisheyeEX method generates more realistic textures and coherent semantics than comparison methods. For better comparisons, we visualize more qualitative extrapolation results as shown in Figure \ref{fig:more_res}.

\noindent \textbf{User Study:} Furthermore, we conduct a user study to compare the photorealism and faithfulness of generated results. Specifically, we randomly select 150 images from three synthesized datasets (50 from Cityscapes, 50 from BDD100k, 50 from KITTI) and select 50 images from the real-world fisheye image dataset LMS, which are further constructed into $\alpha$ group and $\beta$ group, respectively. Subsequently, we invite 20 volunteers with image processing expertise to rank the subjective visual qualities of generated results derived from all comparison methods (RK \cite{17}, HiFill \cite{16}, SRN \cite{13}, Boundless \cite{12}, and Ours). For each selected test image in the $\alpha$ group and $\beta$ group, its six generated results are displayed in random order. Each volunteer is asked to rank the results from the best to the worst, and the counts of the best vote for each method are shown in Figure \ref{fig:user_study}. We can observe that we gain the most votes from 20 volunteers and our FisheyeEX performs better than other methods on both the synthesized and real-world fisheye image datasets.

\subsection{Analysis of Polar Domain}
\label{s4.3}
Having observed that the radial symmetry of the fisheye image, we proposed a novel polar outpainting strategy that transforms the outpainting from the Cartesian domain into the polar domain. Thus, we can reach a more reasonable and effective outpainting direction. Also, the learning path of convolutional kernels keeps consistent with the reconstructed distortion distribution in the polar domain. Such a spiral learning fashion boosts a more accurate and efficient distortion perception. To validate the above descriptions, we conduct two experiments in terms of the distortion perception and image extrapolation on different domains.

\begin{table}
  \caption{Comparison of the network architectures used in Cartesian distortion perception and polar distortion perception.}
  \label{tab:network}
	 \centering
   \begin{tabular}{l|c|c|c|c}
    \hline
    Method & Domain & Backbone & Parameter & Storage \\
    \hline
    \hline
    DDM \cite{8}& Cartesian & Encoder-Decoder & $5.4\times10^7$ & 207M\\
    \hline
    Ours & Polar & Encoder & $1.9\times10^7$ & 76M\\

  \hline
\end{tabular}
\end{table}

\begin{table}
  \caption{Performance comparison of the Cartesian distortion perception and polar distortion perception, where $\mathcal{M}_{rd} = \mathcal{L}_{1} + \mathcal{M}_{hs} + \mathcal{M}_{vs} + \mathcal{M}_{cs}$. The speed of two methods is tested on a NVIDIA GeForce RTX 2080 Ti GPU with a fisheye image.}
  \label{tab:metric}
	 \centering
   \begin{tabular}{l|c|c|c|c|c||c}
    \hline
    Method & $\mathcal{L}_1 \downarrow$ & $\mathcal{M}_{hs} \downarrow$ & $\mathcal{M}_{vs} \downarrow$ & $\mathcal{M}_{cs} \downarrow$ & $\mathcal{M}_{rd} \downarrow$ & Speed $\downarrow$\\
    \hline
    \hline
    DDM& 0.039 &  0.034 &  0.035 &  0.046 &  0.154 & 0.019s\\
    \hline
    Ours & \textbf{0.012} & \textbf{0} & \textbf{0} & \textbf{0} & \textbf{0.012} & \textbf{0.005s}\\

  \hline
\end{tabular}
\end{table}

\noindent \textbf{Cartesian distortion perception vs. Polar distortion perception:} Previous methods learn the distortion of the fisheye image using an encoder-decoder network, estimating the 2D distortion map in the Cartesian domain \cite{8, 11}. However, this manner omits the radial symmetry of the fisheye image. In contrast, we achieve more accurate distortion perception only using a lightweight encoder network in the polar domain. In experiment, we use the network in DDM \cite{8} to conduct the distortion map estimation, and the network comparisons are listed in Table \ref{tab:network}. 

As mentioned in Section \ref{s3.2}, besides the $\mathcal{L}_{1}$ error, we present three new metrics for measuring the radial symmetry of the estimated distortion. Therefore, we can formulate a comprehensive metric $\mathcal{M}_{rd}$ to evaluate the performance of radial distortion perception:  $\mathcal{M}_{rd} = \mathcal{L}_{1} + \mathcal{M}_{hs} + \mathcal{M}_{vs} + \mathcal{M}_{cs}$. The performance comparison of the Cartesian distortion perception and polar distortion perception is shown in Table \ref{tab:metric}. We can observe that our polar perception method significantly outperforms the Cartesian perception method \cite{8} in all evaluation metrics. Note that our method can recover the realistic radial symmetry of the fisheye image, realizing zero error in the horizontal, vertical, and central symmetry evaluations. Besides, due to the lightweight model and straightforward perception, we achieve more accurate estimation with almost 3$\times$ faster than DDM \cite{8}. 

\begin{figure}[t]
\centering
\includegraphics[width=1\linewidth]{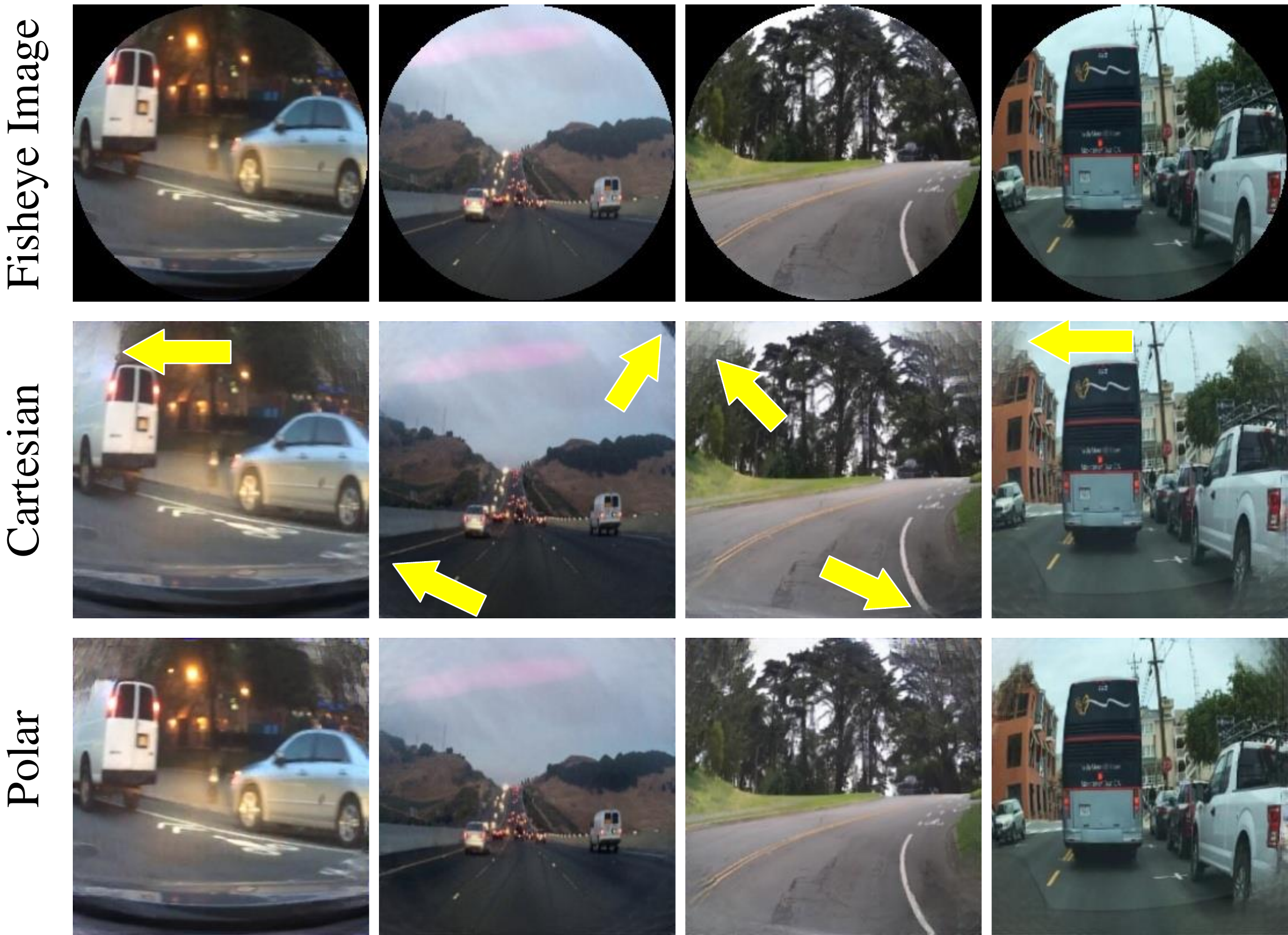}
\caption{Visual comparison of the Cartesian outpainting and polar outpainting. The arrows highlight the unsatisfactory parts.}
\label{fig:cp_cartesian_polar}
\end{figure}

\begin{figure}[t]
\centering
\includegraphics[width=1\linewidth]{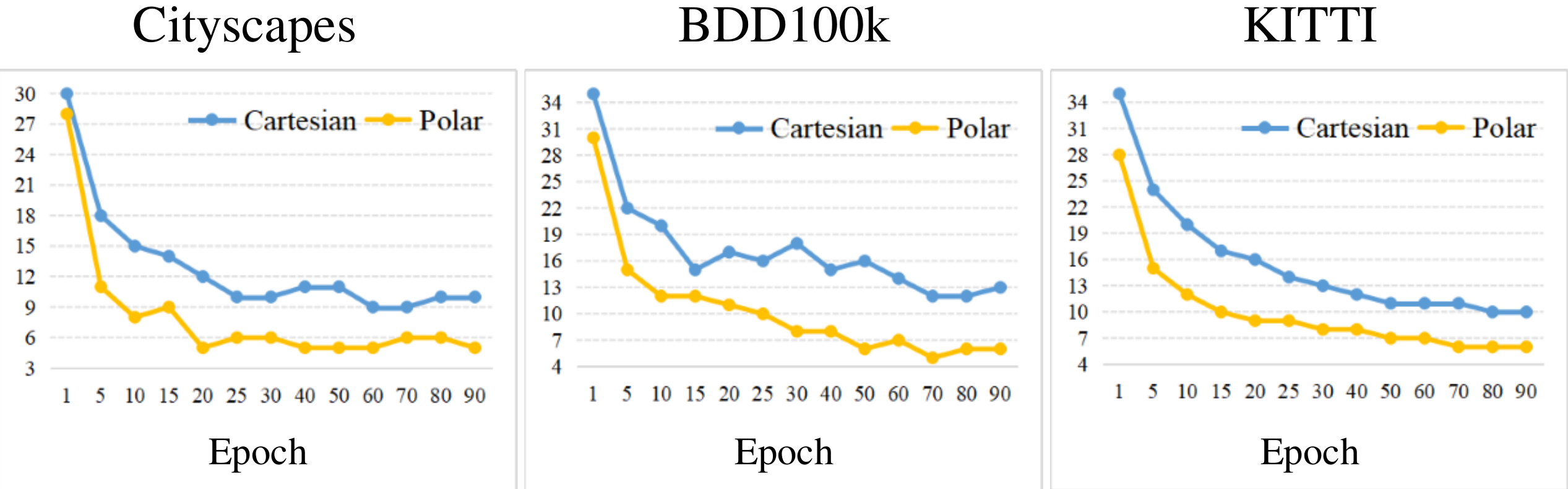}
\caption{Training loss curves of the Cartesian outpainting and polar outpainting.}
\label{fig:loss}
\end{figure}

\noindent \textbf{Cartesian outpainting vs. Polar outpainting:} We also explore the effectiveness of the extrapolation manner for the outpainting performance. For a fair comparison, we only replace our polar outpainting with the conventional Cartesian outpainting, keeping the network configuration the same. Figure \ref{fig:cp_cartesian_polar} shows the visual comparison of the Cartesian and polar outpainting results. With the polar outpainting strategy, our method can reach a reasonable extrapolation direction that generates the plausible content from the center to the outside. By contrast, the outpainting in the Cartesian domain is difficult to hallucinate the semantically consistent details, leading to unsatisfactory outpainting parts, especially in the outside region. For example, at the 1st and 4th columns in Figure \ref{fig:cp_cartesian_polar}, the textures of the car and building are mistakenly borrowed to the outpainting region. Furthermore, to demonstrate the influence of different extrapolation manners on networks' learning ability, we visualize the training loss curves of the Cartesian outpainting and polar outpainting. In Figure \ref{fig:loss}, the polar outpainting manner boosts better and faster convergences than the Cartesian outpainting on all types of datasets, verifying it is more suitable for the fisheye image extrapolation task.

\begin{table}
\begin{center}
\caption{Ablation study of the proposed FisheyeEX method.}
\label{table:ablation}
\begin{tabular}{cccc|ccc}
\hline
  & \ \ Modules&  &  &  &  Metrics &  \\
\hline
 Baseline  & PO & SR &  SDP \ \ & PSNR $\uparrow$ & SSIM $\uparrow$ & FID $\downarrow$\\
\hline
\hline
\cmark & \xmark & \xmark & \xmark & 18.71 & 0.78 & 139.67\\
\cmark & \cmark & \xmark & \xmark & 21.56 & 0.85 & 89.11\\
\cmark & \cmark & \cmark & \xmark & 22.49 & 0.89 & 57.05\\
\cmark & \cmark & \cmark & \cmark & \textbf{23.36} & \textbf{0.92} & \textbf{40.92}\\
\hline
\end{tabular}
\end{center}
\end{table}
\setlength{\tabcolsep}{1.9pt}

\subsection{Ablation Study}
To validate the effectiveness of different components in our method, we conduct an ablation study as reported in Table \ref{table:ablation}. Specifically, we first implement a baseline without the polar outpainting strategy (PO), scene revision module (SR), and spiral distortion-aware perception module (SDP), which conducts the fisheye image extrapolation in the Cartesian domain. Then, we gradually add these removed components to show different performances. From Table \ref{table:ablation}, complete learning with all components equipped achieves the best quantitative evaluation. Additionally, the performance gains improvements as a new component is applied, which verifies the contribution of each component. For example, the polar outpainting strategy significantly improves the extrapolation performance since it enables a more reasonable completion direction, gradually reasoning and generating the coherent semantics from the image center to the outside. Different from the natural image, besides the content generation, the fisheye image outpainting needs to synthesize the radial distortion. Thus, we achieve better outpainting results with the guidance of the scene revision module and spiral distortion-aware perception module. 

\section{Conclusion}
\label{sec5}
This paper introduces the image outpainting into the existing computational photography field, presenting a practical FisheyeEX approach to improve the integrity of captured scenes with larger FoV for the fisheye lens. A novel polar outpainting strategy is proposed to address the specific structure and distortion distribution of the fisheye image in our framework. In the transformed polar domain, we can reach a more effective extrapolation direction that reasons the original content and generates the semantically coherent details from the image center to the outside. Besides the content generation, a spiral distortion-aware perception module can guide accurate radial distortion synthesis. Extensive experiments show that our polar outpainting is superior to previous methods for extrapolating fisheye images. 

However, our solution is difficult to be applied to more complicated cases due to the assumption of the fixed distortion center, it is also a challenging problem in the research field of the distortion rectification and camera calibration. Such a limitation may be addressed by a sequential estimation strategy, which is our future effort direction.

\normalem
\bibliographystyle{ieeetr}

\begin{thebibliography}{100}

  \bibitem{111}
  M.~Sabini and G.~Rusak, ``Painting outside the box: Image outpainting with
    gans,'' {\em arXiv preprint arXiv:1808.08483}, 2018.
  
  \bibitem{12}
  P.~Teterwak, A.~Sarna, D.~Krishnan, A.~Maschinot, D.~Belanger, C.~Liu, and
    W.~T. Freeman, ``Boundless: Generative adversarial networks for image
    extension,'' in {\em Proceedings of the IEEE International Conference on
    Computer Vision}, pp.~10521--10530, 2019.
  
  \bibitem{13}
  Y.~Wang, X.~Tao, X.~Shen, and J.~Jia, ``Wide-context semantic image
    extrapolation,'' in {\em Proceedings of the IEEE Conference on Computer
    Vision and Pattern Recognition}, pp.~1399--1408, 2019.
  
  \bibitem{14}
  Z.~Yang, J.~Dong, P.~Liu, Y.~Yang, and S.~Yan, ``Very long natural scenery
    image prediction by outpainting,'' in {\em Proceedings of the IEEE
    International Conference on Computer Vision}, pp.~10561--10570, 2019.
  
  \bibitem{29}
  D.~Guo, H.~Liu, H.~Zhao, Y.~Cheng, Q.~Song, Z.~Gu, H.~Zheng, and B.~Zheng,
    ``Spiral generative network for image extrapolation,'' in {\em European
    Conference on Computer Vision}, 2020.
  
  \bibitem{15}
  J.~Yu, Z.~Lin, J.~Yang, X.~Shen, X.~Lu, and T.~S. Huang, ``Generative image
    inpainting with contextual attention,'' in {\em Proceedings of the IEEE
    Conference on Computer Vision and Pattern Recognition}, pp.~5505--5514, 2018.
  
  \bibitem{151}
  J.~Yu, Z.~Lin, J.~Yang, X.~Shen, X.~Lu, and T.~S. Huang, ``Free-form image
    inpainting with gated convolution,'' in {\em Proceedings of the IEEE
    International Conference on Computer Vision}, pp.~4471--4480, 2019.
  
  \bibitem{16}
  Z.~Yi, Q.~Tang, S.~Azizi, D.~Jang, and Z.~Xu, ``Contextual residual aggregation
    for ultra high-resolution image inpainting,'' in {\em Proceedings of the
    IEEE/CVF Conference on Computer Vision and Pattern Recognition},
    pp.~7508--7517, 2020.
  
  \bibitem{17}
  L.~Hongyu, J.~Bin, S.~Yibing, H.~Wei, and Y.~Chao, ``Rethinking image
    inpainting via a mutual encoder-decoder with feature equalizations,'' in {\em
    Proceedings of the European Conference on Computer Vision}, 2020.
  
  \bibitem{18}
  C.~Yang, X.~Lu, Z.~Lin, E.~Shechtman, O.~Wang, and H.~Li, ``High-resolution
    image inpainting using multi-scale neural patch synthesis,'' in {\em The IEEE
    Conference on Computer Vision and Pattern Recognition (CVPR)}, July 2017.
  
  \bibitem{181}
  S.~Iizuka, E.~Simo-Serra, and H.~Ishikawa, ``Globally and locally consistent
    image completion,'' {\em ACM Transactions on Graphics (ToG)}, vol.~36, no.~4,
    pp.~1--14, 2017.
  
  \bibitem{19}
  J.~Yu, Z.~Lin, J.~Yang, X.~Shen, X.~Lu, and T.~S. Huang, ``Free-form image
    inpainting with gated convolution,'' in {\em Proceedings of the IEEE
    International Conference on Computer Vision}, pp.~4471--4480, 2019.
  
  \bibitem{20}
  Y.~Zeng, J.~Fu, H.~Chao, and B.~Guo, ``Learning pyramid-context encoder network
    for high-quality image inpainting,'' in {\em The IEEE Conference on Computer
    Vision and Pattern Recognition (CVPR)}, pp.~1486--1494, 2019.
  
  \bibitem{25}
  J.~Kopf, W.~Kienzle, S.~Drucker, and S.~B. Kang, ``Quality prediction for image
    completion,'' {\em ACM Transactions on Graphics (TOG)}, vol.~31, no.~6,
    pp.~1--8, 2012.
  
  \bibitem{26}
  Y.~Zhang, J.~Xiao, J.~Hays, and P.~Tan, ``Framebreak: Dramatic image
    extrapolation by guided shift-maps,'' in {\em Proceedings of the IEEE
    Conference on Computer Vision and Pattern Recognition}, pp.~1171--1178, 2013.
  
  \bibitem{27}
  M.~Wang, Y.-K. Lai, Y.~Liang, R.~R. Martin, and S.-M. Hu, ``Biggerpicture:
    data-driven image extrapolation using graph matching,'' {\em ACM Transactions
    on Graphics}, vol.~33, no.~6, 2014.
  
  \bibitem{28}
  I.~Goodfellow, J.~Pouget-Abadie, M.~Mirza, B.~Xu, D.~Warde-Farley, S.~Ozair,
    A.~Courville, and Y.~Bengio, ``Generative adversarial nets,'' in {\em
    Advances in Neural Information Processing Systems}, pp.~2672--2680, 2014.
  
  \bibitem{1}
  C.~Br\"auer-Burchardt and K.~Voss, ``A new algorithm to correct fish-eye and
    strong wide-angle lens distortion from single images,'' vol.~1, pp.~225--228,
    2001.
  
  \bibitem{2}
  C.~Geyer and K.~Daniilidis, ``Paracatadioptric camera calibration,'' {\em IEEE
    Trans. Pattern Anal. Mach. Intell.}, vol.~24, pp.~687--695, 2002.
  
  \bibitem{3}
  R.~Swaminathan and S.~K. Nayar, ``Nonmetric calibration of wide-angle lenses
    and polycameras,'' {\em IEEE Trans. Pattern Anal. Mach. Intell.}, vol.~22,
    pp.~1172--1178, 2000.
  
  \bibitem{4}
  S.~B. Kang, ``Catadioptric self-calibration,'' in {\em IEEE International
    Conference on Computer Vision}, 2000.
  
  \bibitem{5}
  M.~Alem{\'a}nflores, L.~Alvarez, L.~Gomez, and D.~Santanacedr{\'e}s,
    ``Automatic lens distortion correction using one-parameter division models,''
    {\em Image Processing on Line}, vol.~4, 2014.
  
  \bibitem{6}
  J.~Rong, S.~Huang, Z.~Shang, and X.~Ying, ``Radial lens distortion correction
    using convolutional neural networks trained with synthesized images,'' in
    {\em Asian Conference on Computer Vision}, pp.~35--49, 2016.
  
  \bibitem{7}
  X.~Yin, X.~Wang, J.~Yu, M.~Zhang, P.~Fua, and D.~Tao, ``Fish{E}ye{R}ec{N}et: A
    multi-context collaborative deep network for fisheye image rectification,''
    in {\em European Conference on Computer Vision}, pp.~469--484, 2018.
  
  \bibitem{8}
  K.~Liao, C.~Lin, Y.~Zhao, and M.~Xu, ``Model-free distortion rectification
    framework bridged by distortion distribution map,'' {\em IEEE Transactions on
    Image Processing}, vol.~29, pp.~3707--3718, 2020.
  
  \bibitem{10}
  K.~{Liao}, C.~{Lin}, Y.~{Zhao}, and M.~{Gabbouj}, ``{DR-GAN}: Automatic radial
    distortion rectification using conditional gan in real-time,'' {\em IEEE
    Transactions on Circuits and Systems for Video Technology}, vol.~30, no.~3,
    pp.~725--733, 2020.
  
  \bibitem{11}
  X.~Li, B.~Zhang, P.~V. Sander, and J.~Liao, ``Blind geometric distortion
    correction on images through deep learning,'' in {\em Proceedings of the IEEE
    Conference on Computer Vision and Pattern Recognition}, pp.~4855--4864, 2019.
  
  \bibitem{43}
  O.~Bogdan, V.~Eckstein, F.~Rameau, and J.-C. Bazin, ``Deepcalib: a deep
    learning approach for automatic intrinsic calibration of wide field-of-view
    cameras,'' in {\em Proceedings of the 15th ACM SIGGRAPH European Conference
    on Visual Media Production}, pp.~1--10, 2018.
  
  \bibitem{50}
  K.~Zhao, C.~Lin, K.~Liao, S.~Yang, and Y.~Zhao, ``Revisiting radial distortion
    rectification in polar-coordinate: A new and efficient learning
    perspective,'' {\em IEEE Transactions on Circuits and Systems for Video
    Technology}, 2021.
  
  \bibitem{36}
  K.~Simonyan and A.~Zisserman, ``Very deep convolutional networks for
    large-scale image recognition,'' {\em arXiv preprint arXiv:1409.1556}, 2014.
  
  \bibitem{433}
  M.~Arjovsky, S.~Chintala, and L.~Bottou, ``Wasserstein {GAN},'' {\em arXiv
    preprint arXiv:1701.07875}, 2017.
  
  \bibitem{60}
  C.~Villani, {\em Optimal transport: old and new}, vol.~338.
  \newblock Springer Science \& Business Media, 2008.
  
  \bibitem{46}
  D.~P. Kingma and J.~Ba, ``Adam: A method for stochastic optimization,'' {\em
    arXiv preprint arXiv:1412.6980}, 2014.
  
  \bibitem{44}
  M.~Cordts, M.~Omran, S.~Ramos, T.~Rehfeld, M.~Enzweiler, R.~Benenson,
    U.~Franke, S.~Roth, and B.~Schiele, ``The cityscapes dataset for semantic
    urban scene understanding,'' in {\em Proceedings of the IEEE Conference on
    Computer Vision and Pattern Recognition (CVPR)}, June 2016.
  
  \bibitem{45}
  F.~{Yu}, H.~{Chen}, X.~{Wang}, W.~{Xian}, Y.~{Chen}, F.~{Liu}, V.~{Madhavan},
    and T.~{Darrell}, ``Bdd100k: A diverse driving dataset for heterogeneous
    multitask learning,'' in {\em 2020 IEEE/CVF Conference on Computer Vision and
    Pattern Recognition (CVPR)}, pp.~2633--2642, 2020.
  
  \bibitem{47}
  A.~Geiger, P.~Lenz, and R.~Urtasun, ``Are we ready for autonomous driving? the
    kitti vision benchmark suite,'' in {\em Conference on Computer Vision and
    Pattern Recognition (CVPR)}, 2012.
  
  \bibitem{48}
  A.~Eichenseer and A.~Kaup, ``A data set providing synthetic and real-world
    fisheye video sequences,'' in {\em 2016 IEEE International Conference on
    Acoustics, Speech and Signal Processing (ICASSP)}, pp.~1541--1545, 2016.
  
  \bibitem{49}
  C.~Szegedy, V.~Vanhoucke, S.~Ioffe, J.~Shlens, and Z.~Wojna, ``Rethinking the
    inception architecture for computer vision,'' {\em Proceedings of the IEEE
    Conference on Computer Vision and Pattern Recognition}, pp.~2818--2826, 2016.
  
  \end{thebibliography}

\end{document}